\documentclass[journal,transmag]{IEEEtran}
\usepackage{psfig}
\usepackage{subfigure}
\pdfoutput=1
\usepackage{graphicx}
\usepackage{amsmath}
\usepackage{amssymb}
\usepackage{setspace}
\usepackage{array}
\usepackage{caption}
\usepackage{stfloats}
\usepackage{longtable}
\usepackage{algorithm}
\usepackage{algorithmicx}
\usepackage{algpseudocode}
\usepackage{amsmath}
\usepackage{amsmath}
\usepackage{amssymb}
\usepackage{stfloats}
\usepackage{color}

\floatname{algorithm}{Algorithm}

\makeatletter
\newif\if@restonecol
\makeatother

\makeatletter
\algrenewcommand\ALG@beginalgorithmic{\footnotesize}
\makeatother

\newenvironment{tinyequation}{\begin{equation}\scriptsize}{\end{equation}}

\begin{document}

\title{Estimation of Passenger Route Choice Pattern \\Using Smart Card Data for Complex Metro Systems}

\author{
\IEEEauthorblockN{Juanjuan Zhao,
Fan Zhang,~\IEEEmembership{Member,~IEEE},
Lai Tu,
Chengzhong Xu,~\IEEEmembership{Fellow,~IEEE},\\
Dayong Shen,
Chen~Tian,
Xiang-Yang Li,~\IEEEmembership{Fellow,~IEEE},
 and Zhengxi Li}

\thanks{Copyright (c) 2015 IEEE. Personal use of this material is permitted. However, permission to use this material for any other purposes must be obtained from the IEEE by sending a request to pubs-permissions@ieee.org.
Corresponding author: Fan Zhang (email: zhangfan@siat.ac.cn).

Juanjuan Zhao and Fan Zhang are with Shenzhen Institutes of Advanced Technology, Chinese Academy of Sciences, China and Shenzhen College of Advanced Technology, University of Chinese Academy of Sciences (e-mail: jj.zhao@siat.ac.cn; zhangfan@hust.edu.cn).

Lai Tu is with School of Electronic Information and Communications, Huazhong University of Science and Technology, China (e-mail: tulai.net@gmail.com;).

Chengzhong Xu is with Wayne State University and Shenzhen Institutes of Advanced Technology, Chinese Academy of Sciences, China (e-mail: cz.xu@siat.ac.cn).

Dayong Shen is with Research Center for Computational Experiments and Parallel Systems, National University of Defense Technology, China (e-mail: dayong.shen@nudt.edu.cn).

Chen~Tian is with State Key Laboratory for Novel Software Technology, Nanjing University, China (e-mail: alexandretian@gmail.com).

Xiang-Yang Li is with School of Computer Science and Technology,
University of Science and Technology of China, China (e-mail: xiangyang.li@gmail.com).

Zhengxi Li is with Department of Automation, North China University of Technology, China (e-mail: lizhengxi@ncut.edu.cn).
}
}

\IEEEtitleabstractindextext{%
\begin{abstract}
Nowadays, metro systems play an important role in meeting the urban transportation demand in large cities.
The understanding of passenger route choice is critical for public transit management. The wide deployment of Automated Fare Collection(AFC) systems opens up a new opportunity. However, only each trip's tap-in and tap-out timestamp and stations can be directly obtained from AFC system records; the train and route chosen by a passenger are unknown, which are necessary to solve our problem. While existing methods work well in some specific situations, they don't work for complicated situations. In this paper, we propose a solution that needs no additional equipment or human involvement than the AFC systems. We develop a probabilistic model that can estimate from empirical analysis how the passenger flows are dispatched to different routes and trains. We validate our approach using a large scale data set collected from the Shenzhen metro system. The measured results provide us with useful inputs when building the passenger path choice model.

%
%

\end{abstract}
}

\maketitle

\IEEEdisplaynontitleabstractindextext
\IEEEpeerreviewmaketitle

\begin{figure}[tbp]
\centering
\captionsetup{justification=centering}
\includegraphics[width=0.5\textwidth]{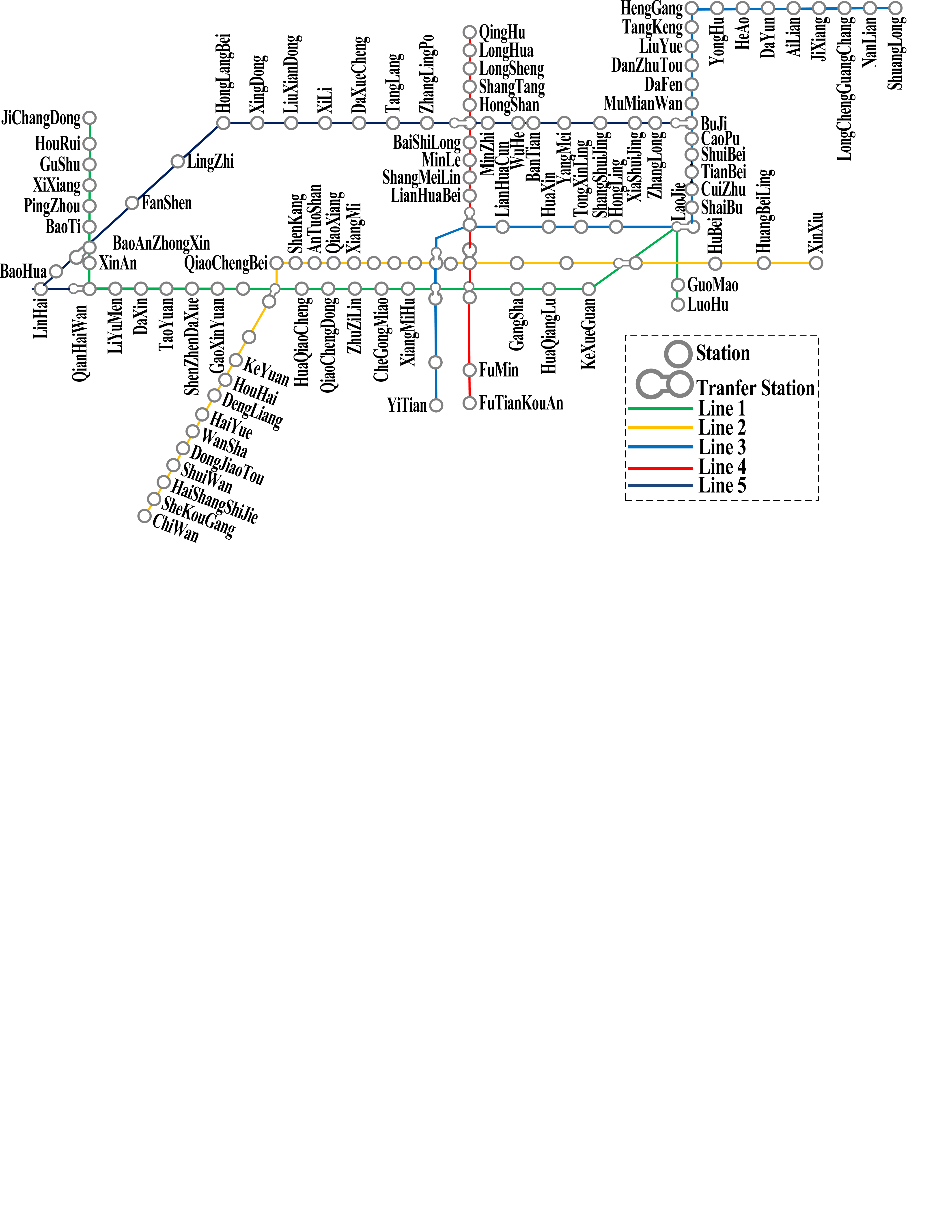}
\caption{Metro graph of Shenzhen}
\label{fig:metro}
\end{figure}

\IEEEraisesectionheading{\section{Introduction}\label{sec:introduction}}
\IEEEPARstart{N}{owadays}, metro systems play an important role in meeting the urban transportation demand in large cities. Due to its fast speed, high efficiency, large volume and punctuality, the urban metro has become the first choice of many people. In Shenzhen, China, in mid-June 2015, there were around 3.5 million metro trips every day, which was around one third of the total public traffic. Fig.~\ref{fig:metro} illustrates the metro operating map of Shenzhen. With further expansion of the metro system, the amount of passengers may increase rapidly. On one hand, the increasing usage of metros can effectively help reduce the traffic pressure on surface roads. On the other hand, it also brings dramatic increasing of passenger demand on metro systems.

\begin{figure*}[tbp]
\centering
\includegraphics[width=1.0\textwidth]{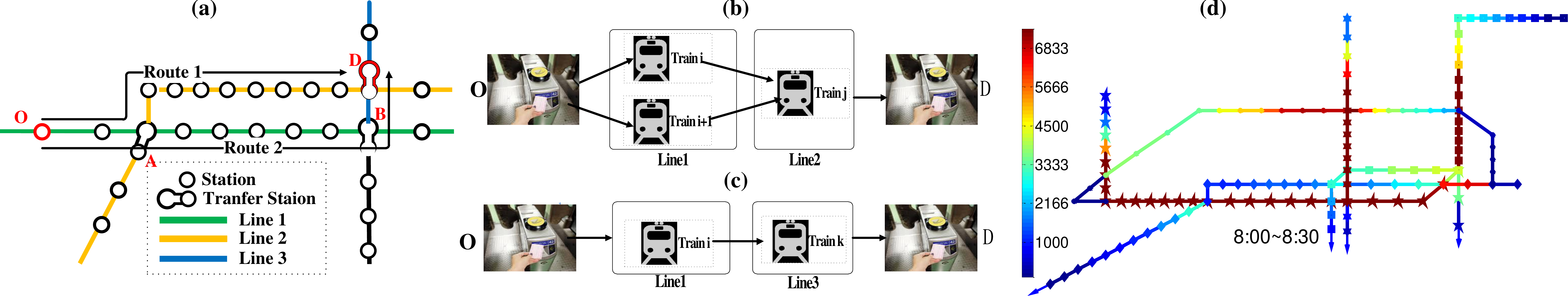}
\caption{(a)An OD with Multiple Routes (b)Trains matching for route 1 (c)Trains matching for route 2 (d)An illustration of traffic monitor application based on the proposed model}
\label{fig:mutilroute}
\end{figure*}

The traffic patterns of large metro systems are usually very complex. Under the condition of network operation and seamless transfer in current metro systems, the train and route chosen by a passenger are unknown. It is common to have more than one route between the origin station and the destination station, a.k.a \emph{multi-path} in transportation systems. As shown in Figure~\ref{fig:mutilroute}(a), there are two routes from station O to station D. This means that for an OD pair with more than one route, we don't know how passengers are distributed over these routes and trains.

This missing information at a fine granularity could be important for both passengers and metro operators. From the operators' point of view, understanding the flow distribution of passengers in the whole metro network is important for improving the service reliability. The potential applications can be a mobile application of trip planning for metro passengers, a monitoring system for metro operators, a route suggestion and emergency management system for urban administrators etc. This paper aims to develop a solution to calculate the probability of each route chosen for an OD pair, which can be used to estimate the passengers flow at a granularity of trains of each line, as shown in Figure~\ref{fig:mutilroute}(d).

Traditional approaches are not scalable. To understand the passengers' route choice behavior, one traditional method is to conduct field surveys at train stations, by asking
passengers which route they will take to reach their destinations. There are limitations of this method: firstly, most surveys are conducted with focus on a part of the passengers at particular locations within a limited time window, hence the results are often limited in diversity, scale and accuracy; secondly, it is both labor-intensive and time-consuming in conducting such surveys.

The wide deployment of Automated Fare Collection(AFC) systems opens up a new opportunity for metro network analysis: the transaction records from AFC can reveal the Origin (O) and the Destination (D) of every passenger's trip, as passengers are required to tap their smart cards or RFID based tickets each time they enter the O station or exit the D station. Passengers' flows can be coarsely demonstrated by OD (origin-destination) pairs. However, AFC records failed to expose the passengers' routes directly. Even in cases that the route of an OD is unique, the AFC records are still not able to show which train a passenger takes. There are too many factors that can affect a passenger's final plan, i.e., trains or train combinations one takes. For example, if the train fails to have enough capacity to accommodate all passengers waiting on a platform, some passengers would have to wait for another train. This phenomenon, known as ``travelers left behind" is quite common during rush hours or at large stations.
There are already some studies using transaction records from AFC to understand the passengers' route and train choice behavior~\cite{kusakabe2010estimation,Sun2015116}. Although these methods work well in some specific situations, they don't work for complicated situations, such as the case where there are various ``left-behinds'' at different stations caused by the imbalance of geographical distribution of passengers. Also, usually the walking time between the charge gate and that platform, and the walking time for transfer between platforms could not be ignored.

%
In this paper, we propose a solution that needs no additional facility than the trains operating time table and the AFC records data. By matching a passenger's smart card records with the trains operating time table, the route that he/she might choose can be narrowed down. We develop a probabilistic model that can empirically estimate how the passenger flows are distributed among different routes and trains. As a concrete example in Figure~\ref{fig:mutilroute}(b)(c), if a passenger taps-in at station $O$ at time point $T1$ and taps-out at station $D$ at time point $T2$, both Route 1 and Route 2 can be the possible choice after we narrow down the possible plan based on the time table. Our solution is to answer that at what probability each route is chosen by the passengers. For a route like Route 1, which further has multiple possible train combinations that satisfy the time table constraints, we further derive the probability that passengers may choose each plan, i.e., $\{tr_i,tr_j\}$ or $\{tr_{i+1},tr_j\}$.

The contributions of this paper include:


\begin{itemize}


\item We define two kinds of time-dependent polynomial distributions of the number of trains waited for by passengers. The first is the number of trains that a passenger waits at his/her original station. The other is the number of trains a passenger waits when he/she transits at the transfer station. A set of algorithms are proposed to calculate the parameters of the two distributions.

\item We further propose a probabilistic model that can estimate how the passenger flows are distributed among different routes and trains.

%
%

\item We then deploy the algorithms on a cloud platform and develop supporting modules for the system level solution.

\item Finally we validate our approach using a large scale data set collected from the Shenzhen metro system. The measured results provide us with useful inputs when we build the passenger path choice model.
\end{itemize}

For the rest of this paper, we discuss the related work in Section~\ref{sec:related}. The overview of this study is given in Sections~\ref{sec:overview}. Section~\ref{sec:Solution} discusses the solution in details. We present system design and the algorithm implementation on a cloud platform in Section~\ref{sec:implement}. Section~\ref{sec:Casestudy} presents the experimental studies. Finally, Section~\ref{sec:Conclusion} concludes the paper.

\section{Related Work}
\label{sec:related}
Building users route choice model is an important research direction in the field of transportation~\cite{DynamicPeeta}, which is the basis for traffic management policies-making. Due to the lack of the observation of how probably each route is chosen for an OD pair with multiple routes, most of the past studies focus on building route choice models from empirical perspective. They assume that all passengers have full knowledge of the transportation when attempting to minimize some objective functions e.g., minimizing their travel time (user equilibrium) or minimizing the total system travel time (system optimum)\cite{UrbanEnglewood,ModelingTalaat,NakayamaUrban}. However, those models depend heavily on behavior assumptions and lack in reliable supporting data. Given the dynamic and stochastic nature of transportation systems, the assumption of the passengers' global knowledge is questionable.

Fortunately, the large amount of smart card data in a long period provide us a great opportunity to analyze passengers' transit behavior and evaluate transit service. There were a few previous studies regarding the utilization of smart card data. The literature~\cite{bagchi2005potential} considered the potential usage of smart card data for travel. The literature~\cite{agard2006mining} analyzed users' travel behavior using data mining technology, which clustered users into four groups according to their temporal travel patterns. Our recent work~\cite{zhao2014understanding} studied individual passenger's temporal and spatial travel patterns. We found that if a passenger is temporally regular, it is very possible that the passenger is also spatially regular. Besides understanding travel behavior, smart card data have been used to improve public transit services. The study in~\cite{pelletier2011smart} gave a comprehensive review of using smart card data from different aspects: strategic, tactical and operational. To improve the resilience to service disruptions of metro systems, paper~\cite{Jin201417} investigated a practical problem about integrating localized bus service with metro network. Using the same data set, three optimization models were formulated to design demand-driven timetables for a single-track metro service~\cite{Sun2014284}.


For understanding passengers' flow assignments in metro system, The authors in~\cite{kusakabe2010estimation} proposed a method to estimate which trains every smart card holder boarded during his/her journey. This method could be used to estimate trains' occupancies. However, it was also based on some assumptions that may be only available in some limited scenarios: (a)The methodology assumed that all passengers know the train timetable beforehand. When choosing route, they will first choose the plan with minimum total waiting time, then choose the plan with fewer transits. The remaining small percentage of undecided trips were assumed to be assigned to all possible plans with an equal probability. (b)The walking time between the charge gates and that platform, and the walking time for transfer between platforms are ignored, which may lead to mismatching between passengers' tap time and trains' operation time.

The authors in~\cite{sun2012using} proposed a linear regression model to analyze the individual trajectory during a metro trip, which could be used to estimate the spatial-temporal density inside a metro system. However, their model focused on a single-track scenario that is oversimplified. The study in ~\cite{FuBayesianroute} used a clustering algorithm to infer the route-use patterns of metro passengers from the smart-card data. It confirmed that a Gaussian mixture model worked well in finding the route shares and the mean and variance of travel times for each route of London underground. But the conclusion based on two preconditions. First, the number of routes used by users must be known in advance. Second, the probability distribution function of travel time of each route must be Gaussian. The study in~\cite{Sun2015116} developed an integrated Bayesian approach to infer both network attributes and passenger route choice behavior in a complex metro system, which worked well in some cases that there are lack of train timetable. But a large set of explanatory variables and the probability distribution of these variables need to be calibrated, such as it assumed that all link costs are characterized by independent normal distributions. This is not always true. Taking the phenomenon ``left-behind" as example, the imbalance of geographical distribution of passengers leads to the various ``left-behinds'' in different stations and results that the time cost does not follow normal distributions in a station for different number of trains that need to be waited for. There are other related work of big data based analysis for smart transportation systems~\cite{tian2014understanding,zhang2014analyzing,huang2014mining,li2014traffic}, while they are not targeting metro systems.

In sum, the existing methods did not consider the passengers' ``left-behind'' in detail, which however is one of the main factors affecting us to understanding passengers' path choice behavior. In this paper, we propose a novel approach to calculate the probability of each route used for an OD with multiple routes, it can be used to complicated traffic situation of complex metro network, especially for the situation that ``left-behind'' occurs often.

\section{Overview}
\label{sec:overview}
\subsection{Dataset}
There are two types of data used in this study, smart card transaction data and train operation data. A smart card transaction record is reported when a passenger passes through the entrance or exit gate by tapping smart card, which includes fields $id$(unique identifier of smart card), $s$(metro station), $t$(transaction time) and $type$(enter or exit). A train operation record is collected when a train arrives at or departs a station, which includes fields $sq$(train sequence), $l$(metro line), $s$(metro station), $t$(transaction time) and $type$(arrive or leave).

For a trip $x$ of a passenger, we can observe the trip's beginning time $x.b$, origin $x.o$, end time $x.e$, and destination $x.d$, by joining the tap-in and tap-out tap events together. If the trip needs $i-1$ transfers, we say the trip has $i$ parts. The first part is from the passenger entering metro system to he/she getting off from his/her first boarded train. The last part is from the passenger getting off from the second last train to he/she exiting metro system. If the passenger doesn't need to transfer, then the first part of the trip is also the last part.

\subsection{Notations and Assumptions}
\label{subsec:assumption}
Suppose the set of effective routes of an OD pair is $R$ and $R=\{R_1,R_2...,R_Z\}$. For simplicity, we divide one day into fixed slots with a time interval $\delta$. We set $\delta$ to be a half hour. Then one day can be split as $I=\{I_1,I_2,I_3,...,I_{48}\}$. And we assume that given the interval $\delta$, the probability of each route being chosen is stable in each time interval. We further define that the routes being chosen in a specific time slot $I_j$ obeys polynomial distribution with parameter $\boldsymbol{\alpha}_j=\{\alpha_{j,1},\alpha_{j,2},...,\alpha_{j,Z}\}$ where $\sum\nolimits_{z = 1}^Z {{\alpha _{j,z}}}  = 1$. Given the train operation table $Tab$, the set of trips of passengers $X=\{x^1,x^2,x^3,...,x^Q\}$ that begin at time slot $I_j$, we aim to calculate $\boldsymbol{\alpha}_j$.

For simplicity, we further present some assumptions and notations that will be used in this paper.
\begin{itemize}
\item We assume that the time that most passengers spend to walk between the platform and the ODs' entrance/exit is less than the departure interval between two adjacent trains. This assumption is rational, because for most metro system, the distance between gate and platform is not far.

\item We assume that most passengers will exit the metro station through the exit gate as soon as possible after getting off the train that reaches her/his destination.
\end{itemize}

Based on the two assumptions, we can infer that given a trip of a passenger and the route that he/she chooses, the train that he/she boards in the last part of trip can be determined uniquely.

\emph{\textbf{Tapin($\zeta$)}}: where $\zeta=(s, l, j)$, represents the passengers who enter metro system at time slot $j$ in station $s$ and chooses metro line $l$.

\emph{\textbf{Transfer($\eta$)}}: where $\eta=(s, l, l', j)$, represents the passengers who transfer from metro line $l$ to $l'$ in transit station $s$ at the time slot $j$.

To calculate $\boldsymbol{\alpha}_j$ of an OD pair, all effective routes are needed firstly. Then given all effective routes $R$, and all trips $X$ starting at time slot $I_j$ from station $O$ to $D$, and train operation table $Tab$, we use the maximum likelihood function as Equation~(\ref{equ:Routefunction}) to calculate $\boldsymbol{\alpha}_j$, where $Pr\left( {{x^q}.e|Tab,{x^q}.b,R_z} \right)$ is the possibility that a passenger ${x^q}$ passes through exit gate at time ${x^q}.e$ on condition of $Tab$, ${x^q}.b$ and the route chosen $R_z$.
\begin{tinyequation}
\begin{split}
L\left( {X,Tab,{\boldsymbol{\alpha} _j}} \right) = \log \prod\limits_{{x^q} \in X} {\left( {\sum\limits_{{R_z} \in R} {\left( {\left( {{\alpha _{j,z}} \times Pr\left( {{x^q}.e|Tab,{x^q}.b,{R_z}} \right)} \right)} \right)} } \right)} \\
 = \log \sum\limits_{{x^q} \in X} {\sum\limits_{{R_z} \in R} {\left( {\left( {{\alpha _{j,z}} \times Pr\left( {{x^q}.e|Tab,{x^q}.b,{R_z}} \right)} \right)} \right)} }
\end{split}
\label{equ:Routefunction}
\end{tinyequation}

In practical, the time cost of a trip ($x.e-x.b$) has a certain relation with train operation data. So given a trip of a passenger, we can find all possible plans (train or trains combination) that the passenger may choose for a route by matching two types of data. So $Pr\left( {{x^q}.e|Tab,{x^q}.b,R_z} \right)$ can be calculated by summing up the probabilities of all plans.

In order to get the probability of each plan being chosen, we first transform the train that a passenger may board into the number of trains being needed to wait for. Then we define that the number of trains waited by passengers of $Tapin(\zeta)$ obeys the polynomial distribution with parameter $\boldsymbol{\theta} _{\zeta}=\{\theta _{\zeta}(1),\theta _{\zeta}(2),...,\theta _{\zeta}(n)\}$, and the number of trains waited by passengers of $Transfer(\eta)$ obeys the polynomial distribution with parameter $\boldsymbol{\beta} _{\eta}=\{\beta _{\eta}(1),\beta _{\eta}(2),...,\beta _{\eta}(n)\}$.

From the process of a trip of a passenger, we can infer that $\boldsymbol{\beta}$ is affected by $\boldsymbol{\theta}$. So we can calculate $\boldsymbol{\theta}$ firstly, then $\boldsymbol{\beta}$. As not all the OD pairs have multiple routes, the trips with one route and no transfer can be used to estimate $\boldsymbol{\theta} _{\zeta}$ because the train chosen is unique. Then considering $\boldsymbol{\theta} _{\zeta}$ as prior knowledge, the trips with one route and some transfers can be used to estimate $\boldsymbol{\beta}$. Finally, considering both $\boldsymbol{\theta} _{\zeta}$ and $\boldsymbol{\beta}$ as prior knowledge, $\alpha_{j,z}$ can be estimated by maximizing function~(\ref{equ:Routefunction}).

\subsection{Framework}
\begin{figure}[tbp]
\centering
\captionsetup{justification=centering}
\includegraphics[width=0.5\textwidth]{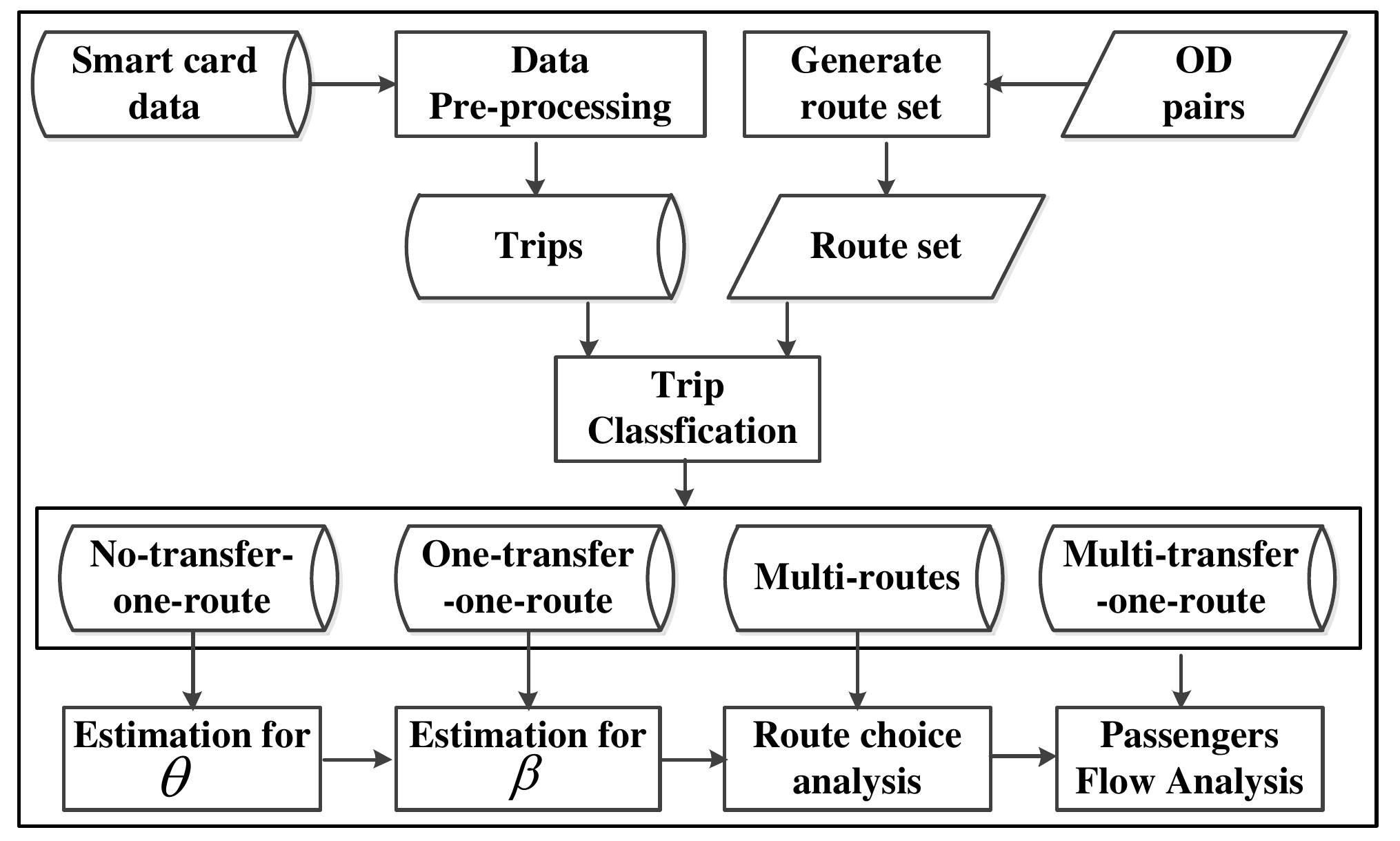}
\caption{Processing flowchart}
\label{fig:processing}
\end{figure}

The framework is illustrated in Figure~\ref{fig:processing}. The details are given as follows.

For smart card data, we have been finding several kinds of errant data, e.g., missing data, duplicated data and data with logical errors. So in the step of data pre-processing, we conduct a detailed clearing process to filter out errant data on a daily basis. In the step of generating route set, we use the algorithm proposed in~\cite{Eppstein} to find the k shortest paths of all OD pairs. Then according to the time cost of passengers in practice, we filter some routes that passengers have never used. In the step of trips classification, according to the number of routes and transfers of their ODs, we classify all trips into four groups: No-transfer-one-route, One-transfer-one-route, Multi-transfer-one-route, Multi-routes. In the step of possible plan analysis, we find all possible plans that a passenger may chose by matching smart card data with trains operation timetable. The trips in No-transfer-one-route and One-transfer-one-route groups are used for estimating $\boldsymbol{\theta}$ and $\boldsymbol{\beta}$ respectively. Then considering $\boldsymbol{\theta}$ and $\boldsymbol{\beta}$ as prior knowledge, we calculate the probability of each route being chosen for an OD with multiple-routes. Finally as an application, passenger flows are analyzed.

\section{Methodology}
\label{sec:Solution}
\subsection{Finding all effective routes for an OD pair}
\label{subsec:effectiveroutes}
In this subsection, we use two steps to find all effective routes for an OD pair. The first step is to find all routes for an OD pair. The second step is to filter the routes that have never been used by passengers from these possible routes. We use the algorithm proposed in~\cite{Eppstein} to find the $k$ shortest routes with efficiency in time $O(m+n\log n+k)$, where $n$, $m$ are the numbers of the vertices and edges in a digraph respectively. We define the cost of a route as the maximum time cost that contains the minimum of walking time and running time of trains. $k$ is determined in term of the accessibility and complexity of metro system. In practice, not all of the $k$ routes of an OD are used by passengers. In order to filter those routes that passengers never choose, we sort all trips of an OD pair over two months by the time cost. We then filter the top $Y\%$ trips with largest time cost. Although most of passengers do not linger too long inside metro system, there are still some passengers showing abnormal travelling behaviors, such as beggars, express logistics worker. Their time cost and travel plan choice may be anomaly. Our recent work~\cite{zhao2014understanding} found that a reasonable value of $Y$ is 2, which can filter the abnormal passengers with high accuracy. Then we get the largest time cost denoted as $C_{max}$ of the remaining $1-Y\%$ trips. Finally we filter the routes with time cost longer than $C_{max}$ from all possible routes. The rest are effective routes. In this paper, if not explicitly pointed out, a route refers to an effective one.

\subsection{Extracting all possible plans chosen by each passenger}
\label{subsec:possibletrains}
\begin{figure}[tbp]
\centering
\captionsetup{justification=centering}
\includegraphics[width=0.5\textwidth]{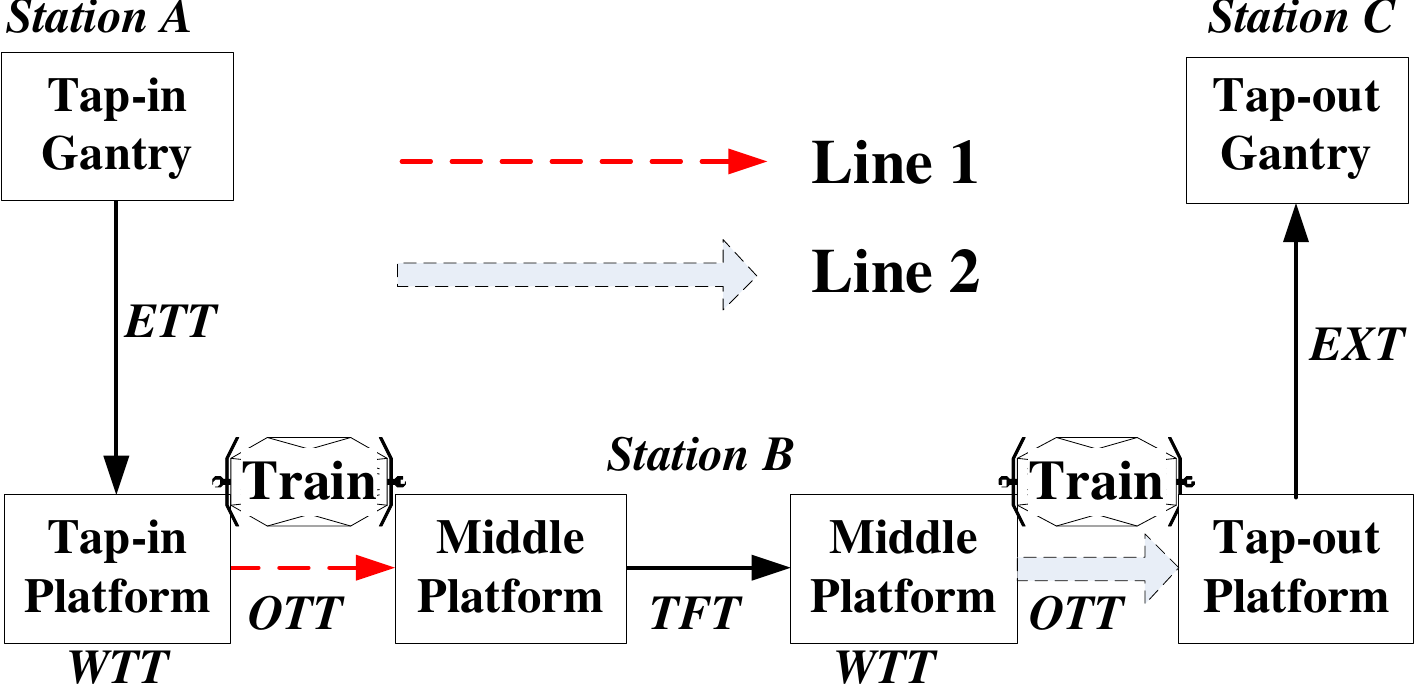}
\caption{Segmentation of a one-transfer trip}
\label{fig:side:trip}
\end{figure}

In this subsection, given a passenger's smart card data and train operation data, we extract all possible plans that can be chosen by the passenger. A general trip of a passenger in metro system can be depicted as 5 steps as shown in Figure~\ref{fig:side:trip}: (1) passing through entrance gate and walking to the platform, (2) waiting on the platform for a train, (3) boarding a train and staying on the train until the train reaches the passenger's destination, (4) getting off the train and exiting the metro system. To be noted, if the passenger needs to transfer, before step(4), (5) transit between platforms needs to be considered. So the whole trip duration is composed of: entry time ({\it ETT}), wait time ({\it WTT}), on train ({\it OTT}), transfer time ({\it TFT}), and exit time ({\it EXT}).

In this paper, we denote the minimal walking time of $ETT$, $TFT$, $EXT$ as $ETT_{min}(l, s)$, $TFT_{min}(l, l', s)$ and $EXT_{min}(l, s)$ respectively, where {\it l} and  {\it l'} are metro lines and {\it s} is a metro station. The method for calculating the value of $TFT_{min}(l, l', s)$ and $EXT_{min}(l, s)$ has been given in our previous paper~\cite{Zhang2015Segmentation}.

Let us denote the arrival and departure time of a train $tr$ at station $s$ of metro line $l$ as $T_{arrv}(l,s,tr)$ and $T_{lv}(l,s,tr)$ respectively. Suppose a passenger $x$ enter metro system at station $s$. His being able to board the train $tr$ needs to satisfy the following Equation~(\ref{equ:condition})(i). Likewise, if a passenger $x$ exits metro system at station $s$, his being able to board the train $tr$ before his exiting metro system needs to satisfy the following Equation~(\ref{equ:condition})(ii). For a passenger who need to transit at transfer station $s$, let us denote the arrival time of a train $tr$ at station $s$ of line $l$ as $T_{arrv}(l,s,tr)$ and the departure time of another train $tr'$ of another line $l'$ at station $s$ as $T_{lv}(l',s,tr')$. That the passenger getting off from $tr$ can board train $tr'$ needs to satisfy Equation~(\ref{equ:condition})(iii).

\begin{equation}
\begin{split}
 &(i) x.b +ETT_{min}(l, s) \leq T_{lv}(l, s, tr). \\
  &(i) x.e-ETE_{min}(l, s) \leq T_{arrv}(l, s, tr). \\
 &(iii)T_{lv}(l', s, tr')-T_{arrv}(l, s, tr) \geq TFT_{min}(l, l', s). \\
\end{split}
\label{equ:condition}
\end{equation}

In sum, for a passenger, given each route, we can find all plans chosen during her/his trip by Equation~(\ref{equ:condition}).

\subsection{Solution of $\boldsymbol{\theta} _{\zeta}$ and $\boldsymbol{\beta} _{\eta}$}
In this section, we give the approaches for calculating $\boldsymbol{\theta} _{\zeta}$ and $\boldsymbol{\beta} _{\eta}$. As aforementioned, we define that the number of trains waited by the passengers of $Tapin(\zeta)$ obeys the polynomial distribution with parameter $\boldsymbol{\theta _{\zeta}}=\{\theta _{\zeta}(1),\theta _{\zeta}(2),...,\theta _{\zeta}(n)\}$, and the number of trains waited by passengers of $Transfer(\eta)$ obeys the polynomial distribution with parameter $\boldsymbol{\beta} _{\eta}=\{\beta _{\eta}(1),\beta _{\eta}(2),...,\beta _{\eta}(n)\}$. We consider the two polynomial distributions $\boldsymbol{\theta} _{\zeta}$ and $\boldsymbol{\beta} _{\eta}$ separately. That's because the transfer passengers arrive at transit station almost simultaneously. While the time that the passengers arrive at the origin station is more random. Hence we first solve that given a plan chosen by a passenger, how to transform it into the number of trains that the passenger waits for. Then we give an approach to estimate $\boldsymbol{\theta} _{\zeta}$ and $\boldsymbol{\beta} _{\eta}$ using several specific trips.

\subsubsection{The number of trains waited by passengers}
\begin{figure}[tbp]
\centering
\captionsetup{justification=centering}
\includegraphics[width=0.5\textwidth]{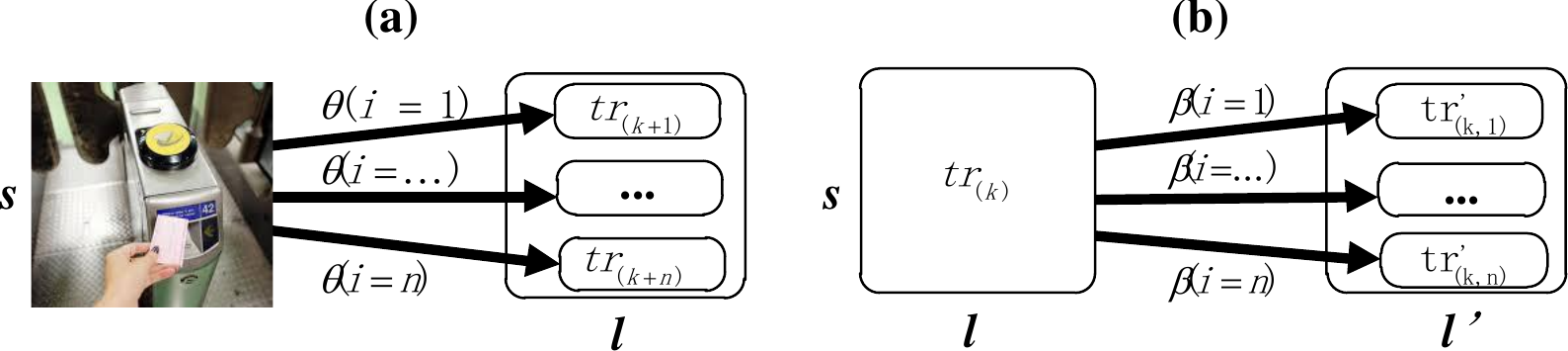}
\caption{The number of trains waited by passengers (a) of $\textit{Tapin}(\zeta)$ \ \  (b) of $\textit{Transfer}(\eta)$ }
\label{fig:thetabeta}
\end{figure}

Given a train boarded by a passengers of $Tapin(\zeta)$, in order to transform it into the number of trains the passenger waited for, we divide these passengers of $Tapin(\zeta)$ into several groups according to the arrival time of trains. We use $tapin(\zeta,k)$ of $Tapin(\zeta)$ to represent the passengers who enter the metro system between the departure time of train $tr_{(k)}$ and the departure time of train $tr_{(k+1)}$. Suppose for these passengers in $tapin(\zeta,k)$, the set of trains that they may board is $\{tr_{(k+1)},tr_{(k+2)},...,tr_{(k+n)}\}$, as shown in Figure~\ref{fig:thetabeta}(a). $n$ is the maximum number of trains needed to be waited for. Field observations show that the first-come-first-served policy is not applicable in practice. There are many factors affecting which train a passenger eventually gets on, such as the distance between the gate and the platform, walking speed, the number of passengers in the waiting queue. Furthermore, a typical train has six to eight cars with multiple doors available for boarding simultaneously. A wise strategy or good luck in choosing train doors could also lead to an earlier boarding. So the train that a passenger eventually gets on is more likely to be a random variable in practice. Let the probability of train $tr_{(k+i)}$ boarded by these passengers is $\theta _{\zeta}(i)$. Thus the number of trains needed to be waited for (the number of passengers in these trains) obeys to polynomial distribution, where $\sum_{{i=1}}^n {\theta _{\zeta}(i)}=1$.

Likewise, given a train boarded by the passengers of $Transfer(\eta)$, in order to transform it into the number of trains the passengers waited for, we divide these passengers of $Transfer(\eta)$ into several groups according to the arrival time of trains of line $l$. We use $transfer(\eta,k)$ to represent the passengers who get off train $tr_{(k)}$ of line $l$. Suppose for these passengers in $transfer(\eta,k)$, the set of trains that they transfer is $\{tr'_{(k,1)},tr'_{(k,2)},\cdots,tr'_{(k,n)}\}$ as shown in Figure~\ref{fig:thetabeta}(b). For a passenger, the train that he may get on is also influenced by many factors, such as the distance between platforms, walking speed, the waiting position for train, the number of people in the waiting queue, and so forth. So we see the train that a passenger eventually gets on as a random variable. We assume the probability of the number of trains needed to be waited for is stable in same time slot. We denote the probability of $tr'_{(k,i)}$ boarded by a passengers of $transfer(\eta,k)$  as $\beta _{\eta}^i$. Thus the number of passenger in these trains obeys to polynomial distribution, where $\sum_{i=1}^n {\beta_{\eta}^i}=1$.


\subsubsection{Calculating of $\boldsymbol{\theta} _{\zeta}$ and $\boldsymbol{\beta} _{\eta}$}
From the process of trips of a passengers, we can infer that $\boldsymbol{\beta} _{\eta}$ is affected by $\theta _\zeta(i)$ ($\boldsymbol{\theta} \rightarrow \boldsymbol{\beta}$). So we can estimate $\boldsymbol{\theta} _{\zeta}$ first, then for $\boldsymbol{\beta} _{\eta}$.

In order to calculate $\theta _{\zeta}(i)$, we assume that several specific trips of $Tapin(\zeta)$ are representative enough to analyze the distribution of the number of trains needed to wait for at an origin station. This is a practical because it is difficult to ascertain the exact train chosen for every passenger during the first part of his trip, especially for complex scene with multiple transfers and multiple routes. However the train chosen for a trip with only one route and no transfer can be inferred. So we classified the trips with only one route as group 1. According to the our assumption in Section~\ref{subsec:assumption}, we can know that given the route chosen, the train chosen in the last part of a trip can be determined uniquely. That means the train boarded at the first part of the trip of group 1 can be inferred, because these trips only have one part during their total journeys. And according to our statistics, the volume of trips in group 1 accounts for $30\%$ of the whole. Clearly, those trip are representative enough~\cite{SamplingStatistics}. So we can use these trips with only one route and no transfer needed to estimate $\boldsymbol{\theta}$.

Similarly we assume several specific trips of $Transfer(\eta)$ are representative enough to analyze the distribution of the number of trains needed to wait for at transfer station when passengers need to transit. A trip with only one route and one transfer has two parts during its journey. We classified these trips as group 2 in this paper. The train boarded in the last part of trips in group 2 can be inferred uniquely. Though there may be more than one trains passengers boarded in the first part, the possibility for these trains can be known from $\boldsymbol{\theta}$. So we can see $\boldsymbol{\theta}$ as prior knowledge and use trips in group 2 to estimate $\beta _{\eta}(i)$.

So, we divide all trips into four classes according to the number of routes and transfers of their ODs: No-transfer-one-route, one-transfer-one-route, multi-transfer-one-route, Multi-routes. The passengers in No-transfer-one-route and One-transfer-one-route are used for estimating $\boldsymbol{\theta}$ and $\boldsymbol{\beta}$ respectively. Then using the passengers in Multi-routes and considering $\boldsymbol{\theta}$ and $\boldsymbol{\beta}$ as prior knowledge, we calculate the probability of each route being chosen for an OD with Multi-routes in the next subsection.

Suppose the set of passengers of $Tapin(\zeta)$ in group 1 (one-route-no-transfer) is $X=\{x^{(1)},x^{(2)},x^{(3)},\cdots,x^{(Q)}\}$. Using the approach given in Section~\ref{subsec:possibletrains}, we can get the number of trains that each passenger waits for in $X$ is $W=\{w^{(1)},w^{(2)},w^{(3)},\cdots,w^{(Q)}\}$ respectively. Suppose the number of passengers waiting $i$ trains is $c_i$ by counting the same digits of $W$. We use maximum-likelihood estimation to obtain the value of $\boldsymbol{\theta} _{\zeta}$ by Equation~(\ref{equ:forthetal1}) and (\ref{equ:forthetal2}).

\begin{subequations}
\begin{equation}
L(X,Tt,{\boldsymbol{\theta} _\zeta }) = \log \sum\limits_{i = 1}^n {{{\left( {{\boldsymbol{\theta} _\zeta }(i)} \right)}^{{c_{\rm{i}}}}}}
 \label{equ:forthetal1}
\end{equation}
\begin{equation}
{\boldsymbol{\theta} _\zeta }(i) = \frac{{{c_{\rm{i}}}}}{{\sum\limits_{j = 1}^n {{c_j}} }}
 \label{equ:forthetal2}
\end{equation}
\end{subequations}


Suppose the set of passengers of $Transfer(\eta)$ in group 2 is $X=\{x^{(1)},x^{(2)},x^{(3)},...,x^{(Q)}\}$. They may has different original stations and beginning time slots. We use $\zeta^{(q)}$ to represent the first part of the trip of passenger $x^{(q)}$. Given a passenger $x^{(q)}$, suppose the set of plans that the passenger may choose is $p_{q}=\{p_{q}^{1},p_{q}^{2},...,p_{q}^{M^q}\}$. The train chosen in the second part of trip can be obtained. A plan $p_{q}^{m}$ can be represented as $\{tr_{q}^{m,1},tr_{q}^{2}\}$. The numbers of trains needed to be waited for are $\{w_{q}^{m,1},w_{q}^{m,2}\}$. We use maximum-likelihood estimation to obtain the value of $\boldsymbol{\beta} _{\eta}$ by function Equation~(\ref{equ:betaestimate}). It is difficult to calculate the derivatives of the logarithm of the sum of some formulas in Equation~(\ref{equ:betaestimate1})for maximum. So firstly we convert it to Equation~(\ref{equ:betaestimate2}) by applying the Jensen inequality, and then calculate $\boldsymbol{\beta}$ by maximizing the value of right hand side of Equation~(\ref{equ:betaestimate2}).


\begin{subequations}
\begin{equation}
\begin{split}
L(X,Tt,\boldsymbol{\theta} ,{\boldsymbol{\beta} _\eta }) = \log \prod\nolimits_{q = 0}^Q {Pr\left( {{p_q}|{x^q}.b,\boldsymbol{\theta} ,{\boldsymbol{\beta} _\eta }} \right)} \\
 = \sum\limits_{q = 0}^Q {\log \left( {\sum\limits_{m = 1}^{{M^q}} {Pr(w_q^{m,1},w_q^{m,2}|{x^q}.b,\boldsymbol{\theta} ,{\boldsymbol{\beta} _\eta }} } \right)} \\
 = \sum\limits_{q = 0}^Q {\log \left( {\sum\limits_{m = 1}^{{M^q}} {{\boldsymbol{\theta} _{{\zeta ^{(q)}}}}(w_q^{m,1}) \times {\boldsymbol{\beta} _\eta }(w_q^{m,2})} } \right)}
  \label{equ:betaestimate1}
  \end{split}
\end{equation}
\begin{equation}
 \ge \sum\limits_{q = 0}^Q {\left( {\sum\limits_{m = 1}^{{M^q}} {{\boldsymbol{\theta} _{{\zeta ^{(q)}}}}(w_q^{m,1})\log {\boldsymbol{\beta} _\eta }(w_q^{m,2})} } \right)}
 \label{equ:betaestimate2}
\end{equation}
 \label{equ:betaestimate}
\end{subequations}

\subsection{Calculating the probability of each route being chosen for an OD pair}
In this subsection, we aim to give an approach to calculate the probability of each route being chosen for an OD pair with multiple effective routes. Suppose the set of effective routes from station $O$ to $D$ is $R=\{R_1,R_2...,R_Z\}$. We denote the probability of route $R_z$ being chosen at time slot $I_j$ is denoted as $\alpha_{j,z}$, where $\sum\nolimits_{z = 0}^Z {{\alpha _{j,z}}}  = 1$.

Suppose the set of the passengers entering metro system during time slot $I_j$ from station $O$ to station $D$ is $X=\{x^1,x^2,x^3,...,x^Q\}$, where $Q$ is the number of passengers. We assume that they are independent. For a passenger $x^q$ in $X$, given that he chooses route $R_z$, we can obtain all plans that the passenger may choose during his trip using the approach given in section~\ref{subsec:possibletrains}. Denote the set of plans as $p_{q,z}$, where $Pr\left( {{p_{q,z}}|{x^q}.b,R_z,\boldsymbol{\theta} ,\boldsymbol{\beta}} \right)$ is the possibility that a passenger ${x^q}$ choose $p_{q,z}$ on condition of ${x^q}.b$, route chosen $R_z$, $\boldsymbol{\theta}$ and $\boldsymbol{\beta}$. For the same reason, it is difficult to calculate the derivatives of the logarithm of the sum of some formulas in Equation~(\ref{equ:lkhfunction1}), so we transform it into Equation~(\ref{equ:lkhfunction2}).


\begin{subequations}
\begin{equation}
\begin{split}
L\left( {X,Tt,\boldsymbol{\theta} ,\boldsymbol{\beta} ,{\boldsymbol{\alpha} _j}} \right) = \\
\sum\limits_{{x^q} \in X} {\log \sum\limits_{{R_z} \in R} {\left( {{\alpha _{j,z}} \times Pr\left( {{p_{q,z}}|{x^q}.b,R_z,\boldsymbol{\theta} ,\boldsymbol{\beta} } \right)} \right)} }
 \label{equ:lkhfunction1}
 \end{split}
\end{equation}
\begin{equation}
 \ge \sum\limits_{{x^q} \in X} {\sum\limits_{{R_z} \in R} {\left( {{\alpha _{j,z}} \times \log Pr\left( {{p_{q,z}}|{x^q}.b,R_z,\boldsymbol{\theta} ,\boldsymbol{\beta} } \right)} \right)} }
 \label{equ:lkhfunction2}
\end{equation}
 \label{equ:lkhfunction}
\end{subequations}

\section{System Implementation}
\label{sec:implement}
\begin{figure}[tbp]
\centering
\captionsetup{justification=centering}
\includegraphics[width=0.5\textwidth]{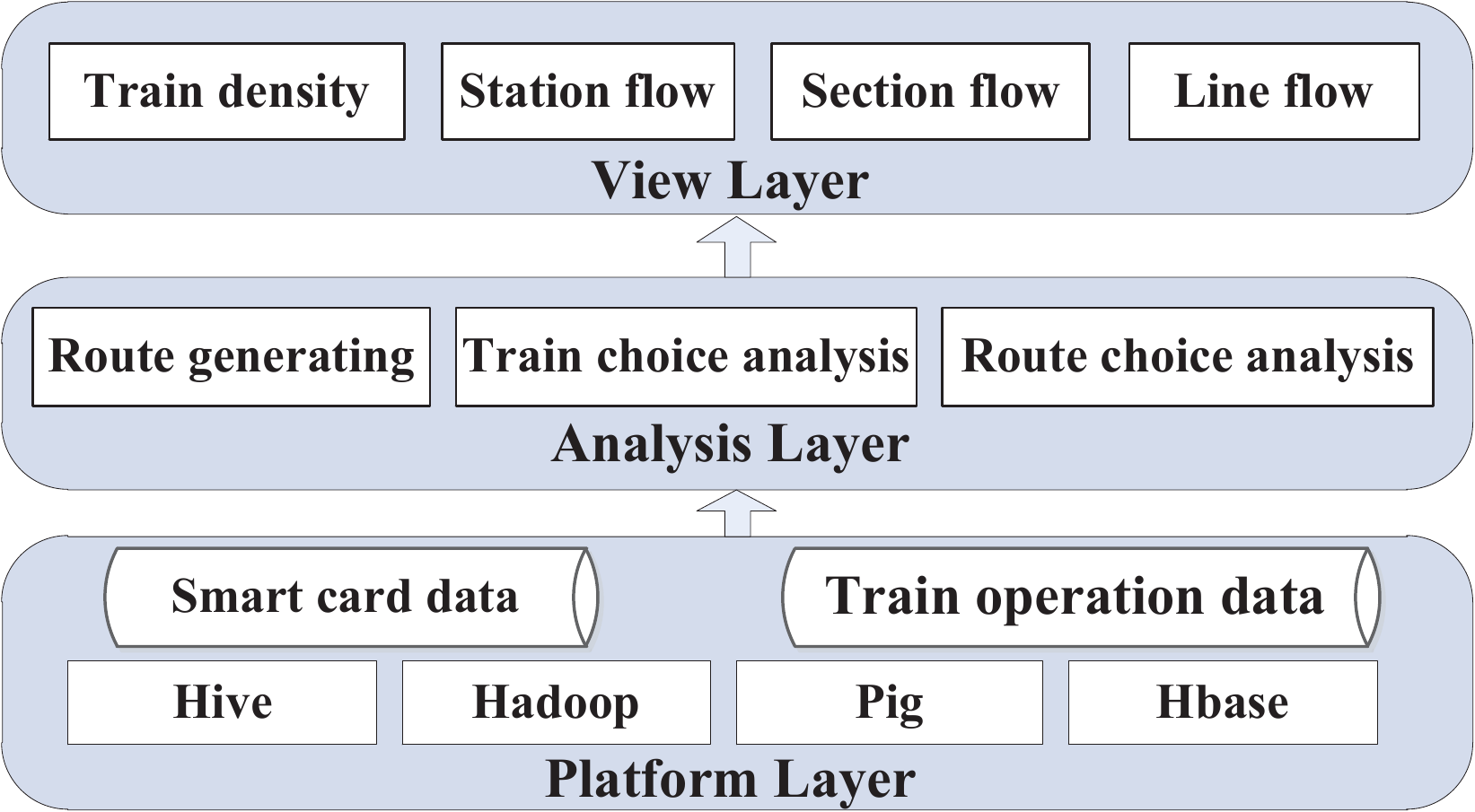}
\caption{System architecture}
\label{fig:system}
\end{figure}

Our algorithm of calculating the probability of each route being chosen for an OD pair with multiple routes is based on a large amount of data. The framework of our system is illustrated in Figure~\ref{fig:system}, which has three layers: the platform layer, the analysis layer and the view layer.

The platform layer is mainly used for storage and job processing purpose. Our algorithms need batch processing on a large amount of data. So it is more efficient to run on a parallel platform~\cite{wang2010parallel}. We use distributed computing platform Hadoop~\cite{white2012hadoop} that was designed for batch processing in big data. It mainly includes two modules, HDFS~\cite{borthakur2008hdfs} and MapReduce~\cite{dean2008mapreduce}. HDFS provides high-throughput access to large data. MapReduce is for parallel processing of large data sets. In our platform, we utilize a 34 TB Hadoop Distributed File System (HDFS) on a cluster consisting of 11 nodes, each of which is equipped with 32 cores and 32 GB RAM. To improve retrieving efficiency, some mapReduce based software tools, such as Pig~\cite{gates2009building}, Hbase~\cite{george2011hbase} are used.

The analysis layer running on platform layer is the keystone of our paper. It mainly contains three parts: Route generating for getting all effective plans of each OD pair; Train choice analysis for finding all possible trains that a passenger may get on; Route choice analysis for calculating the probability of each route being chosen for an OD pair with multiple routes. The three parts are based on large volume of data. They are all being translated to a series of MapReduce jobs that run on the distributed environment.

The view layer based on analysis layer performs passenger flow analysis and displays the results to public or transport agencies for strategic planning and management, such as the spatio-temporal passenger flow analysis for all metro lines, trains, sections, and so on.

%
%
%

\section{Case study}
\label{sec:Casestudy}
\subsection{Dataset}
The dataset used in this study is the smart card transaction
records and train operation logs in Shenzhen, China. The metro system has 5 metro
lines by 2013. The whole data collected from around 4 million smart cards have more than 300 million smart card transaction records, covering 60 consecutive days from June 1, 2015 to July 30, 2015.

\subsection{Left behind analysis}
\begin{figure}[tbp]
\centering
\captionsetup{justification=centering}
\includegraphics[width=0.5\textwidth]{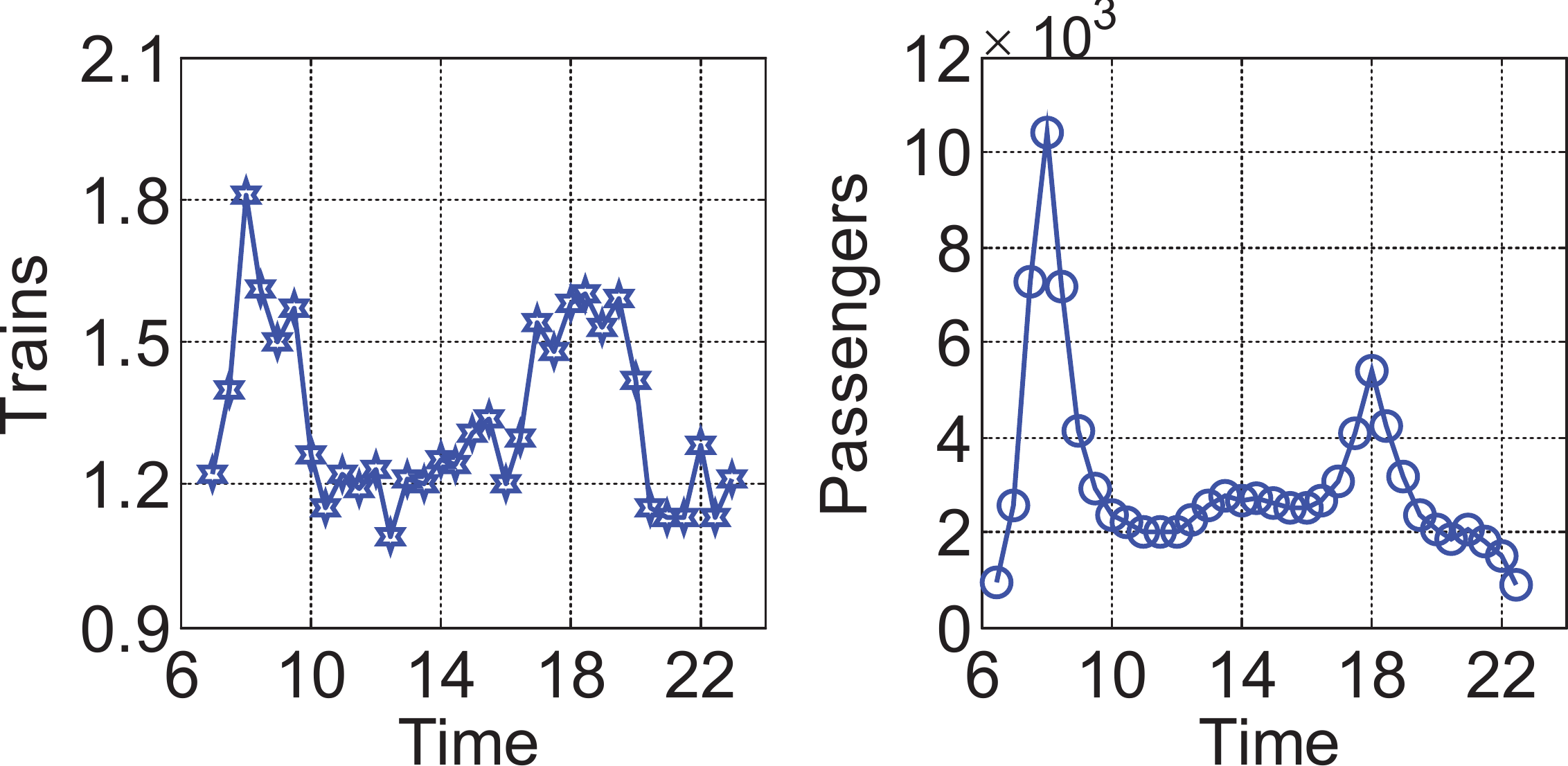}
\caption{(a)Average number of trains waited (b)Sectional flow(LaoJie $\sim$ DaJuYuan)}
\label{fig:boardingpropallday}
\end{figure}

Figure~\ref{fig:boardingpropallday}(a) shows the average number of trains waited by passengers in station \textit{LaoJie} of metro line(\textit{LuoHu}$\sim$\textit{JiChangDong}) at the first part of their trips at different time slots of one day. Figure~\ref{fig:boardingpropallday}(b) shows the number of passengers passing through station \textit{LaoJie} (sectional flows of two adjacent sites \textit{LaoJie} to \textit{DaJuYuan}) at different time slots of one day. The station \textit{Laojie} locating in the heart of ShenZhen business district is a transfer station of line 1 and line 3. There are about 12 thousand tap-in passengers and 60 thousand transfer passengers in \textit{LaoJie} per day. From figure~\ref{fig:boardingpropallday}, we can get that there are a remarkable similarity of the two lines. The cross-correlation of the number of train waited and sectional flow is 0.75 which is larger than 0.7. So we can assume that the more the passengers passing through a station, the more left behind passengers there are in the station.

From Figure~\ref{fig:boardingpropallday}(a) we also can get that the phenomena of left behind is varying with time, which is a good indicator of transit service performance and can provide better travel advice for users. There are two obvious peak periods 7:00$\sim$9:00 and 18:00$\sim$20:00 in Figure~\ref{fig:boardingpropallday}. That is because there are so many passengers who go to work in the morning rush hours and back home in the evening rush hours that the capacity of trains can not meet the actual requirements. So in rush hours, passengers must wait for more trains. Another point that deserves further explanation is that even during off-peak periods that train may not be crowded, the average number of trains needed to be waited for is bigger than 1.0. That is, not all of passengers get on the first available train. This is understandable because some passengers care about comfort that they anticipate that the next train will have seats available and choose to wait.

\begin{figure*}[tbp]
\centering
\includegraphics[width=1.0\textwidth]{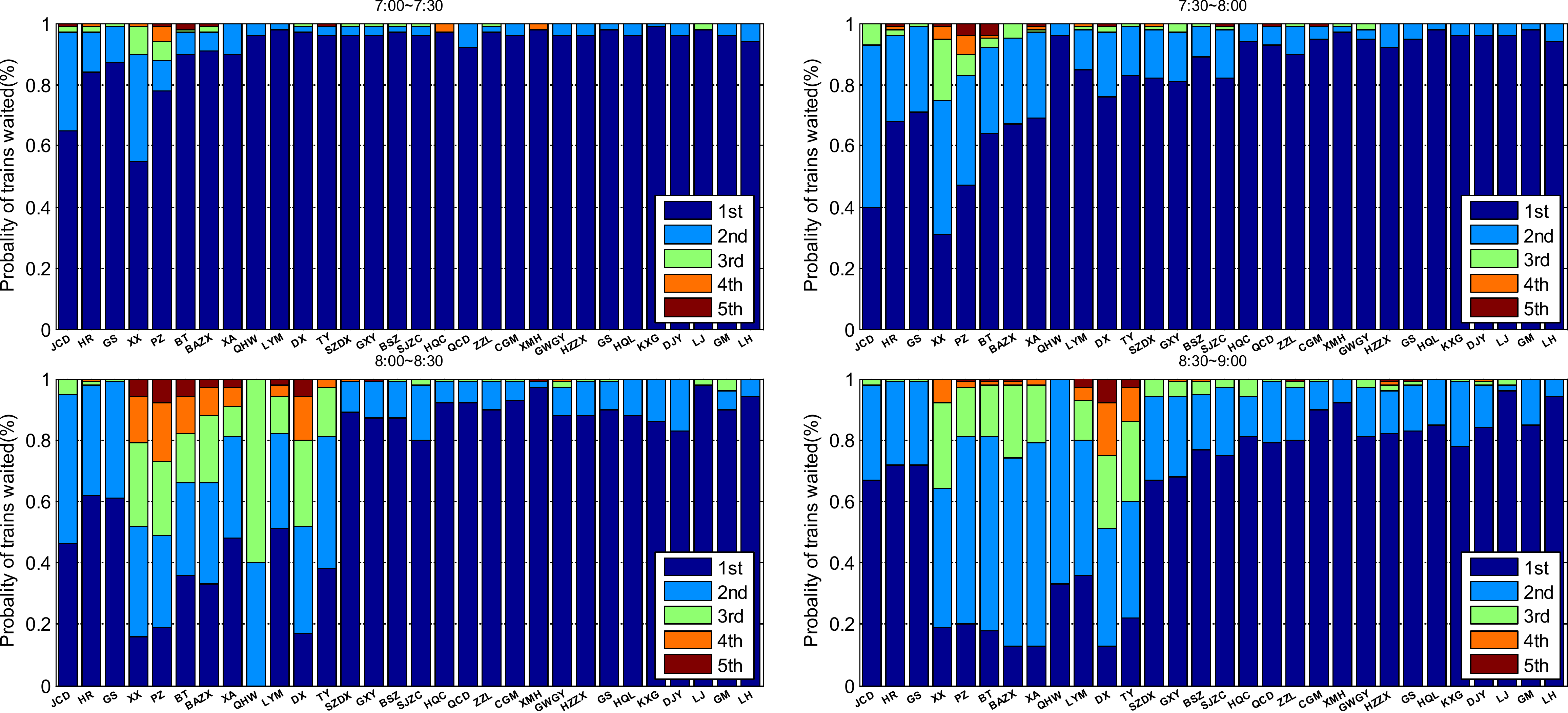}
\caption{The probability of the number of trains needed to wait of line 1(from JCD to Luohu) at four time slots in AM peek}
\label{fig:boardingprop}
\end{figure*}

\begin{figure}[tbp]
\centering
\captionsetup{justification=centering}
\includegraphics[width=0.5\textwidth]{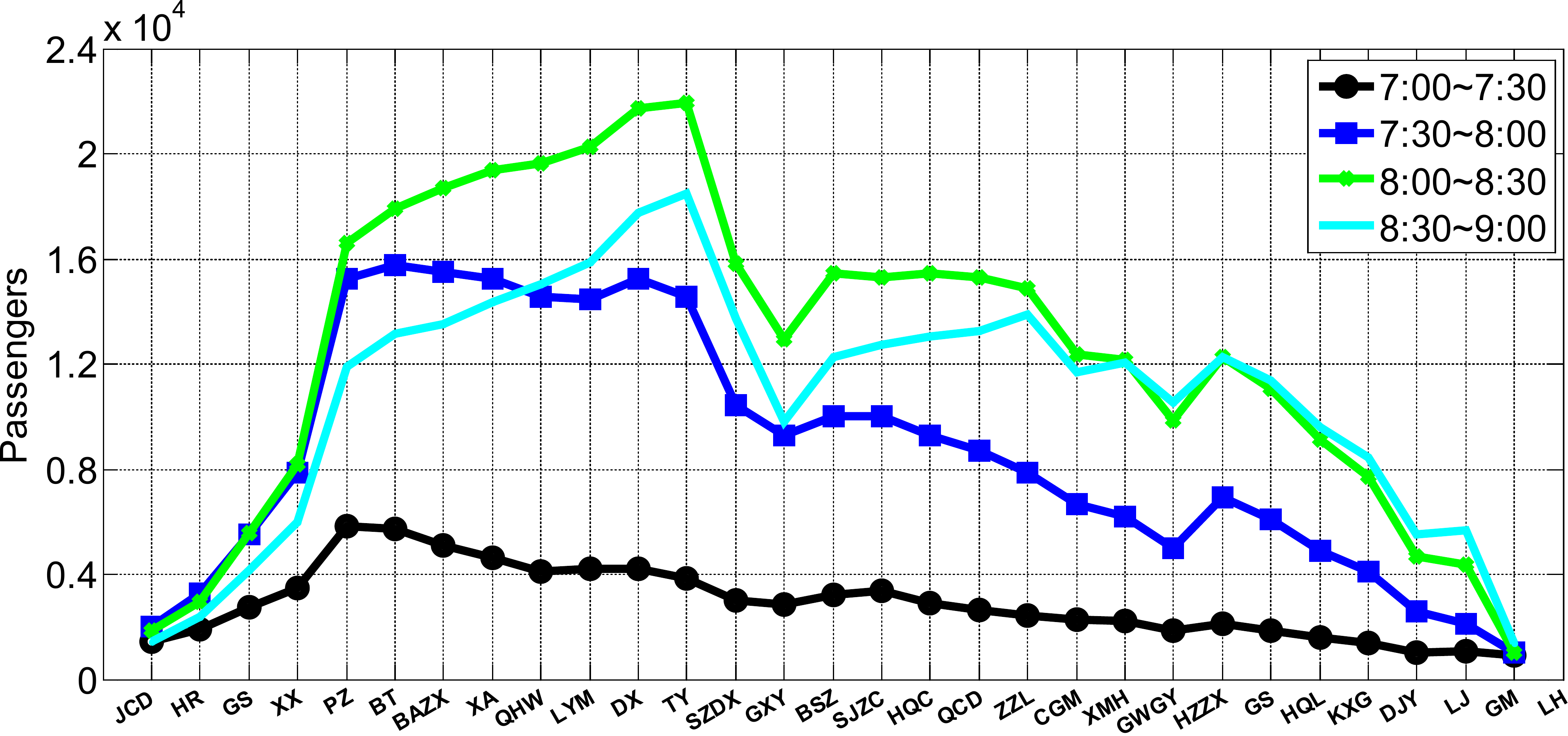}
\caption{Sectional flows of line 1(from JCD to Luohu) at four time slots in AM peek}
\label{fig:sectionflowallstation}
\end{figure}

Figure~\ref{fig:boardingprop} gives the distribution of the number of trains waited by passengers at all stations of metro line 1(\textit{JiChangDong-LuoHu}) at four time slots(07:00$\sim$07:30, 07:30$\sim$08:00, 08:00$\sim$ 08:30, 08:30$\sim$ 09:00) of morning rush hours. In Figure~\ref{fig:boardingprop}, the bar labelled with ``\emph{1st}'' means the probability that a passenger needs to wait for one train (gets on the 1st coming train). ``\emph{2nd}'' means the probability that a passenger needs to wait for two trains, and so on. From Figure~\ref{fig:boardingprop}, we can get that the left-behind varies in time and space. That is because the distribution of passengers is uneven in time and space as shown in Figure~\ref{fig:sectionflowallstation}.

Figure~\ref{fig:sectionflowallstation} gives all sectional flows of metro line 1(\textit{JiChangDong-LuoHu}) at four time slots(07:00$\sim$07:30, 07:30$\sim$08:00, 08:00$\sim$08:30, 08:30$\sim$09:00) of morning peek. We can get that the left-behind is most serious from \emph{XX}(short of \emph{\textit{XiXiang}}) to \emph{\textit{TY}} (short of \emph{\textit{TaoYuan}}) in 08:30$\sim$08:30, which indicates that the train capacity can not meet the demand of passengers in these station.

Moreover, station \emph{\textit{JCD}}(\textit{JiChangDong}) is the first station of line 1 from \emph{\textit{JCD}} to \emph{\textit{Luohu}}, which tell us that all trains arriving at this station are empty. That means that there are more remaining capacity than other stations. But from this figure we get that there are also many passengers in \emph{\textit{JCD}} who need to wait for more than one trains. There are two reasons: Firstly, \emph{\textit{JCD}} is the closest metro station to ShenZhen airport, where there are a lot of passengers getting off from plane and carry packs of luggage and struggle for a local train. Secondly, for safety and comport, they are more likely to choose the next train with more seats. As \emph{\textit{JCD}} is the first station of line 1, passengers prefer a train with seats more than other stations.

\subsection{Route choice pattern}
\begin{figure*}[tbp]
\centering
\includegraphics[width=1.0\textwidth]{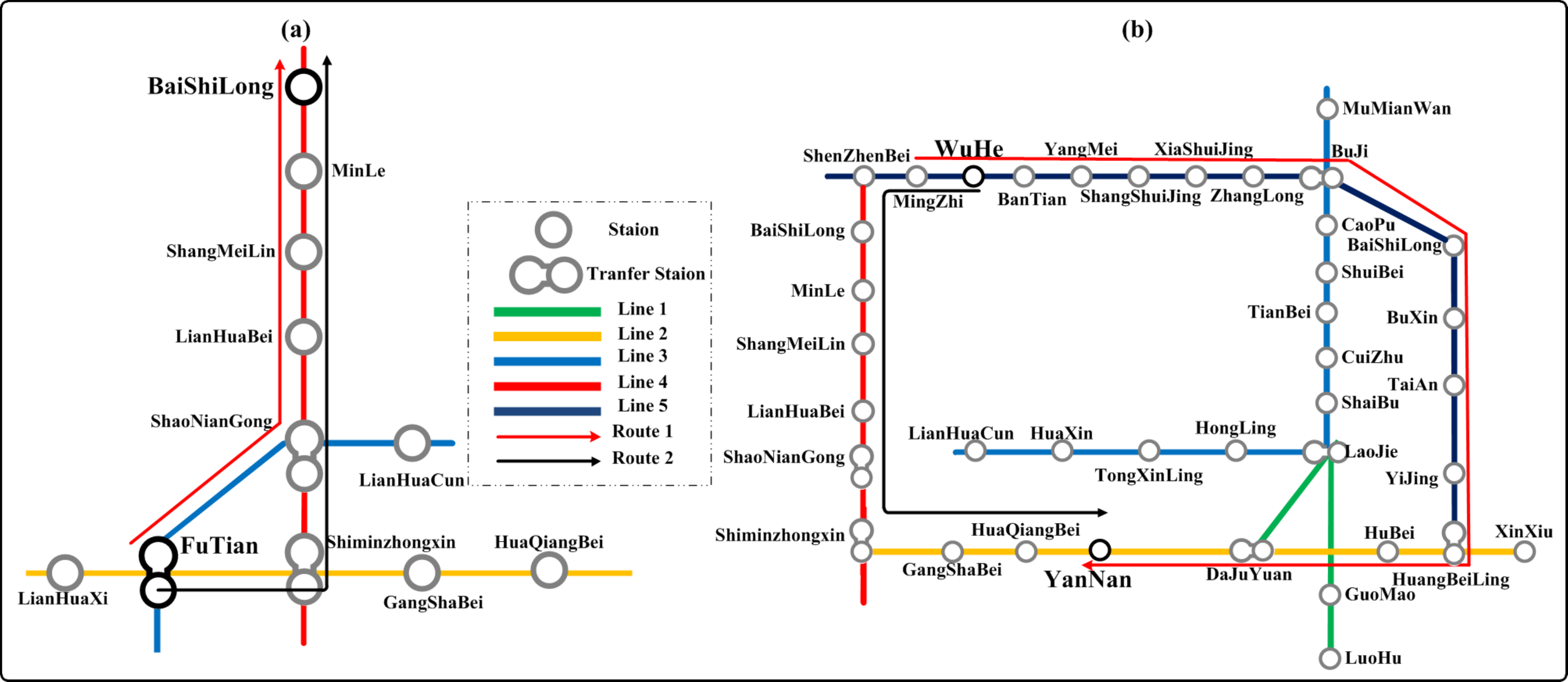}
\caption{ Layout of two OD pairs}
\label{fig:routeanalysis}
\end{figure*}

In this section, two typical OD pairs of stations were chosen to illustrate our proposed method to calculate the probability of each route being chosen. Figure~\ref{fig:routeanalysis}(a) shows the layout of the two OD pairs.

The first OD pair is \emph{BaiShiLong-FuTian} that had two effective routes, (1)Take the metro line 4, get off at station \emph{ShaoNianGong}. Then take metro line 3, get off at the \emph{FuTian} station. (2)Take the metro line 4, get off at \emph{ShiMinZhongXin}. Then take metro line 2, get off at the \emph{FuTian} station. Both of the two routes need one transfer and average time cost of them is nearly the same. The first route is recommended by some mobile apps, such as Baidu map App, ShenZhen metro App provided by ShenZhen Metro Group Company.  However our experiment results show that the probability of the first route being chosen at rush hours and off peak hours is $31\%$ and $42\%$ respectively, which is less than the probability of the second route being chosen. The results tell us that the route given by mobile app doesn't always reflect most passengers' real choice. It also provides proof of the walking penalty when the general cost of a path is calculated.

\begin{figure*}[tbp]
\centering
\includegraphics[width=1.0\textwidth]{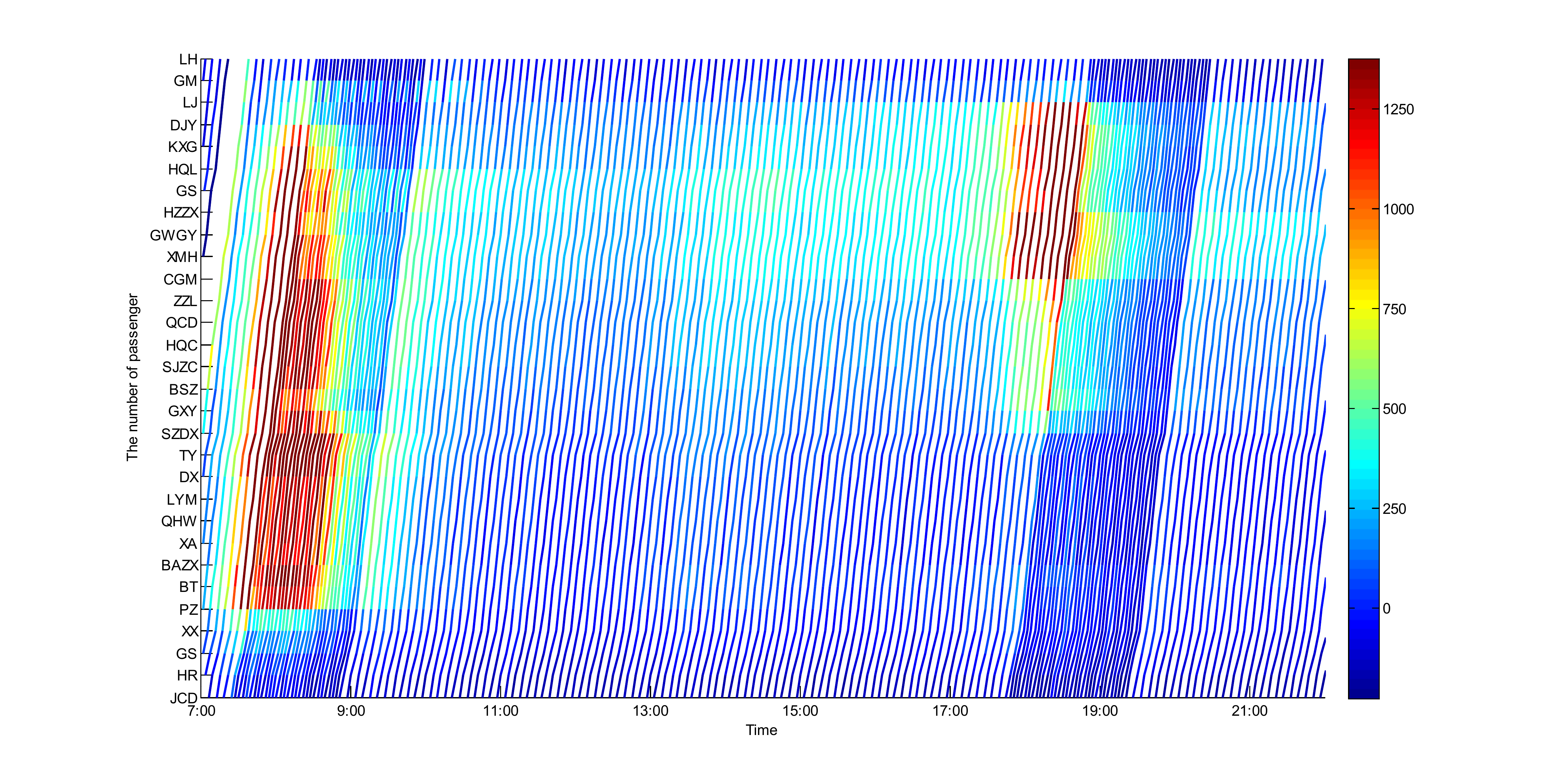}
\caption{Spatio-temporal density of all trains}
\label{fig:density}
\end{figure*}
 \begin{figure*}[tbp]
\centering
\includegraphics[width=1.0\textwidth]{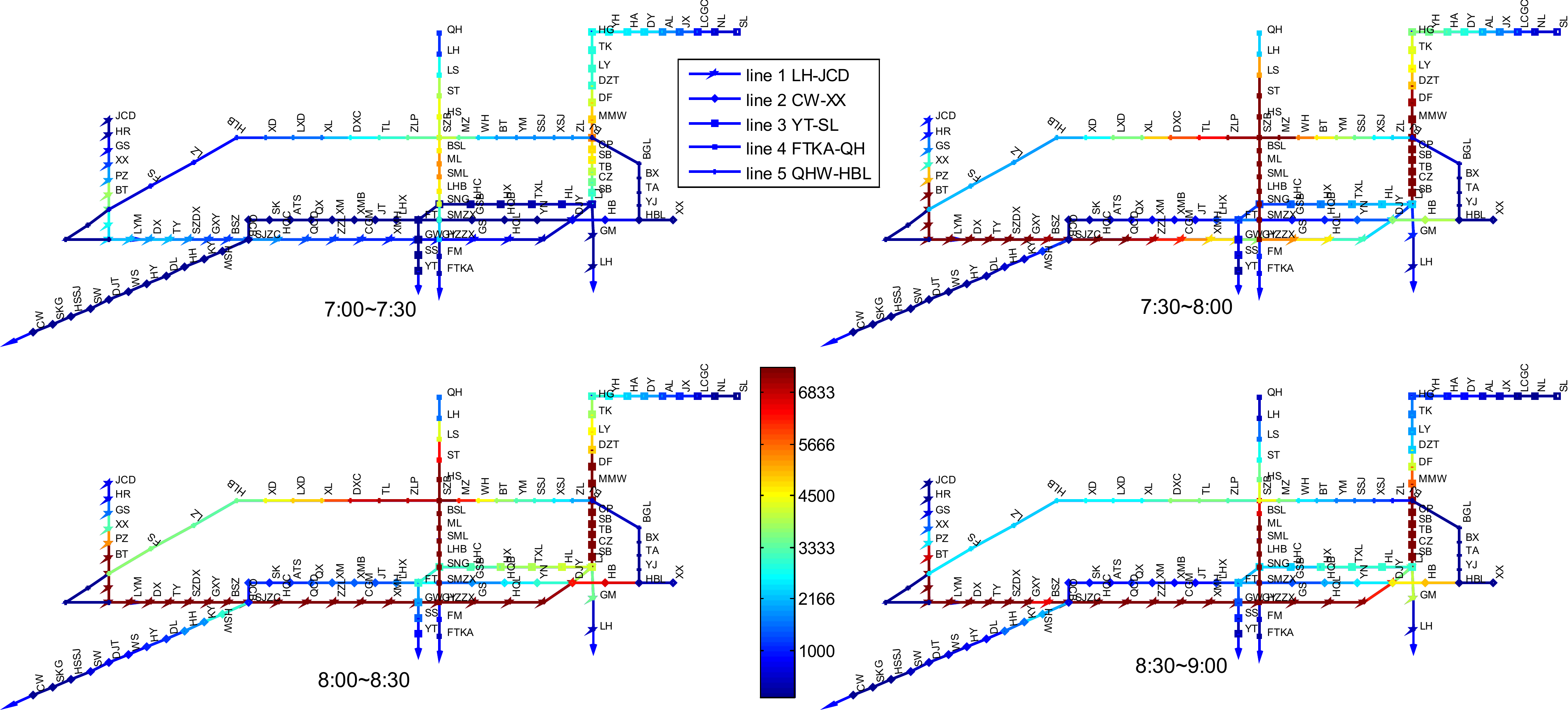}
\caption{Metro sectional flow at AM peek hours}
\label{fig:sectionflow}
\end{figure*}
According to a survey about all transfer stations in ShenZhen, \emph{ShaoNianGong} is one of the transfer stations with longest walking time. The walking time is about 5 minute, which is more than that of \emph{ShiMinZhongXin} with 2 minute. Our on-site investigations tell us that most of passengers do not prefer to walking two much time when transferring. Some passengers tell us that they do not know the actual walking time in every transfer station. Generally, they will follow the shortest path by some map app in their smartphone, which tell them the first route has less time cost than the second one. Based on that, it is understandable for the passengers' route choice behavior of \emph{BaiShiLong-FuTian}. Most of passengers at peak period are local residents. They are more familiar with the metro and know more about the walking time cost than the passengers in off peak period such as visitors. Tourists who have less experience are more likely to rely on mobile apps. So they are more likely to choose to transfer in \emph{ShiMinZhongXin}.

The second OD pair is \emph{WuHe-YanNan}. There are also two effective routes as shown in Figure~\ref{fig:routeanalysis}(b). (1)Take metro line 5, get off at \emph{HuangBeiLing}. Then take metro line 2. (2)Take the metro line 5, get off at \emph{ShenZhenBei}, then take metro line 4, get off at \emph{ShiMinZhongXin}, then take metro line 2. The first route costs ten minutes more than the second one. However our analysis results show that there are still $40\%$ passengers choosing the first route. This is because the second route has one more transfers, which is likely to offset the advantage of low time cost. The result provides proof of the transfer penalty when the general cost of a path is calculated.

\subsection{Spatio-temporal density analysis }
Spatio-temporal density of all trains of metro line 1(Luohu-Jichangdong) is shown in Figure~\ref{fig:density}, where the x axis and y axis represent time and station respectively. Every train starts at the lowest station and finally reaches the highest station in the y axis. Each diagonal line represents a train and covers the information about the train' spatio-temporal density. The color represents the density of passengers. From this figure we can get that there are two peek hours as morning and evening. The density in morning peek is more serious than in evening peek. So the departure intervals of trains in morning peek could be set to be shorter than that in evening peek.

Beside, this spatio-temporal density information can be used for assessing the metro service and forecasting the density of all running train and so on. The sectional flow of whole metro system at four time slots in morning peek is shown in Figure~\ref{fig:sectionflow}. We can get that (1) The passenger flows are most crowded in 8:00$\sim$8:30 (2) The passenger flow is uneven distributed throughout whole metro system. The metro line 1, 3, 4 have more passengers than line 2 and 5. For metro line 1, 3, 4, the densities are more serious in middle than that in both sides. This can help to design schedules for shuttle trains and so on. All these information can improve both passengers' and transportation operators' knowledge on transportation system. For example the information can be further used to improve the service by redesigning timetable, adjusting velocity, and etc. Passengers on the other hand can also plan their trips based on the information.

\section{Conclusion and discussion}
\label{sec:Conclusion}
In this paper, we present an approach to calculate the probability of route choices for an OD pair with multiple routes in a complex metro network. In doing so, we find, for each passenger, all possible plans that he/she can choose for each effective route by matching smart card data and train operation logs. We also calculate two kinds of time-dependent polynomial distributions using maximum likelihood estimation. One is the number of trains that a passenger waits for at his/her original station. The other is the number of trains that a passenger waits for when he/she changes at the transfer station. Based on that, we propose a probability model to calculate the probability of each route being chosen for an OD with multi-paths. The approach in this paper is applied to Shenzhen metro system. On-site investigations validate that our algorithm is accurate and can be used to estimate passenger flows.

\section*{Acknowledgment}
The authors would like to thank anonymous reviewers for their valuable comments.
This work was supported in part by the China National Basic Research Program (973 Program) under Grant 2015CB352400,
by the National High Technology Research and Development Program of China (863 Program) under Grant 2014AA01A702,
by NSFC under Grant U1401258,
by NSF under Grant CMMI 1436786 and CNS 1526638,
by Research Program of Shenzhen under Grant JSGG20150512145714248, KQCX2015040111035011 and CYZZ20150403111012661,
by the Natural Science Foundation of Hubei Province under Grant 2014CFB1007,
and by the Fundamental Research Funds for the Central Universities.


\bibliography{main}
\bibliographystyle{IEEEtran}
\begin{IEEEbiography}[{\includegraphics[width=1in,height=1.25in,clip,keepaspectratio]{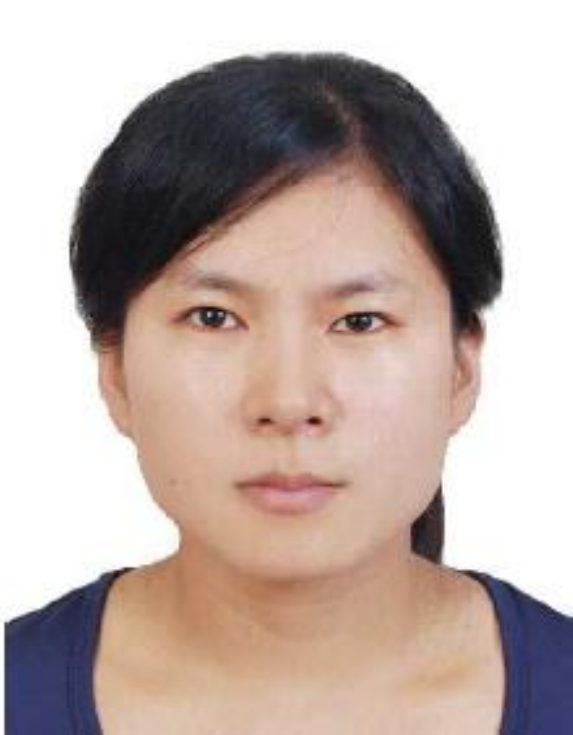}}]{Juanjuan Zhao}
received her MS from the Department of Computer Science from Wuhan University of Technology, China in 2009. She was a research assistant from 2009 to 2012 in the Shenzhen Institutes of Advanced Technology, Chinese Academy of Sciences and currently is a Ph.D. student there. Her research interests include cloud computing, big data processing, streaming-data processing, data fusion technique, big-data-driven systems, spatio-temporal data mining.
\end{IEEEbiography}

\begin{IEEEbiography}[{\includegraphics[width=1in,height=1.25in,clip,keepaspectratio]{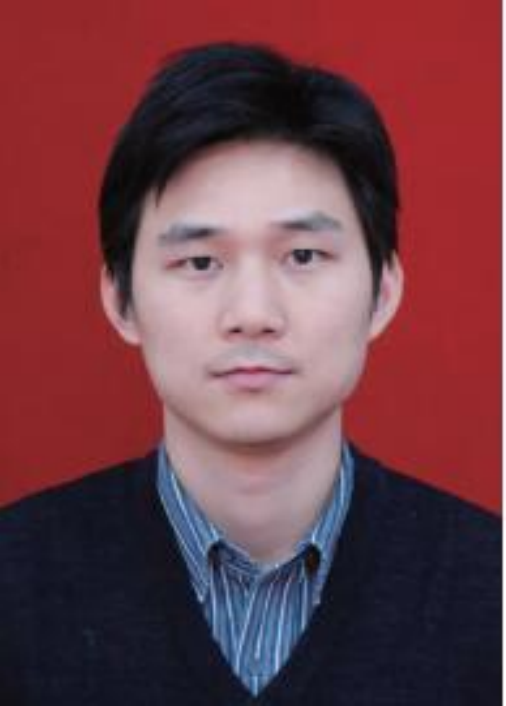}}]{Fan Zhang}
is an associate professor in the Shenzhen Institutes of Advanced Technology, Chinese Academy of Sciences.
He received his Ph.D. in Communication and Information System from Huazhong
University of Science and Technology in 2007. He was a postdoctoral fellow
at University of New Mexico and University of Nebraska-Lincoln from 2009 to 2011. His research topics include big data processing, data privacy and urban computing.
\end{IEEEbiography}

\begin{IEEEbiography}[{\includegraphics[width=1in,height=1.25in,clip,keepaspectratio]{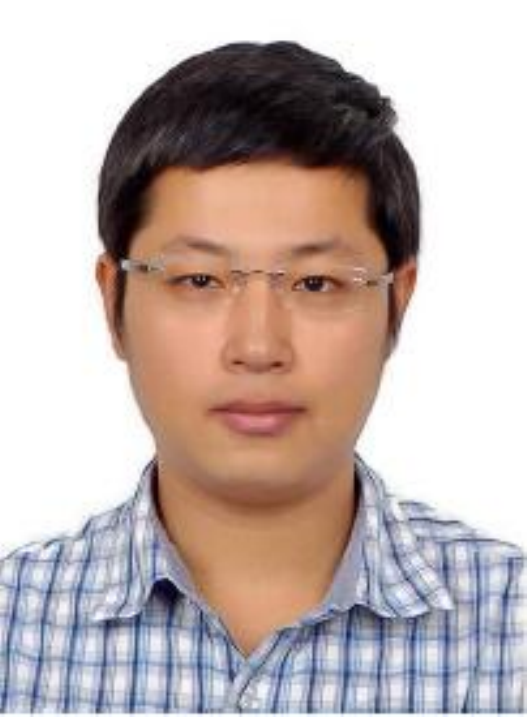}}]{Lai Tu}
received the B.S in Communication Engineering and Ph.D. degree in Information and Communication Engineering from Huazhong University of Science and Technology, China, in 2002 and 2007 respectively. From 2007/7 to 2008/12, he worked as a postdoc fellow in the Department of EIE. in Huazhong University of Science and Technology. From 2009/1 to 2010/10, He worked as a postdoc researcher in the Department of CSIE. in Nation Cheng Kung University, Taiwan. Currently, he is an associate professor of the School of Electronic and Information and Communications in Huazhong University of Science and Technology. His research areas include urban computing, human behavior study, mobile computing and networking.
\end{IEEEbiography}

\begin{IEEEbiography}[{\includegraphics[width=1in,height=1.25in,clip,keepaspectratio]{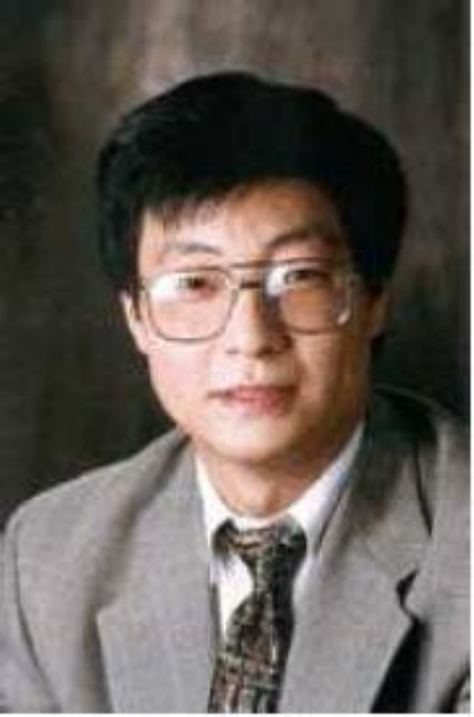}}]{Chengzhong Xu} received his Ph.D. degree from the University of Hong Kong in 1993. He is currently a professor of the Department of Electrical and Computer Engineering of Wayne State University, USA. He also holds an adjunct appointment with the Shenzhen Institute of Advanced Technology of Chinese Academy of Science as the Director of the Institute of Advanced Computing and Data Engineering.  His research interest is in parallel and distributed systems and cloud computing. He has published more than 200 papers in journals and conferences. He was the Best Paper Nominee of 2013 IEEE High Performance Computer Architecture (HPCA), and the Best Paper Nominee of 2013 ACM High Performance Distributed Computing (HPDC). He serves on a number of journal editorial boards, including IEEE Transactions on Computers, IEEE Transactions on Parallel and Distributed Systems, IEEE Transactions on Cloud Computing, Journal of Parallel and Distributed Computing and China Science Information Sciences. He was a recipient of the Faculty Research Award, Career Development Chair Award, and the President¡¯s Award for Excellence in Teaching of WSU. He was also a recipient of the ``Outstanding Oversea Scholar¡± award of NSFC.  For more information, visit http://www.ece.eng.wayne.edu/~czxu. Dr. Xu is an IEEE Fellow.
\end{IEEEbiography}

\begin{IEEEbiography}[{\includegraphics[width=1in,height=1.25in,clip,keepaspectratio]{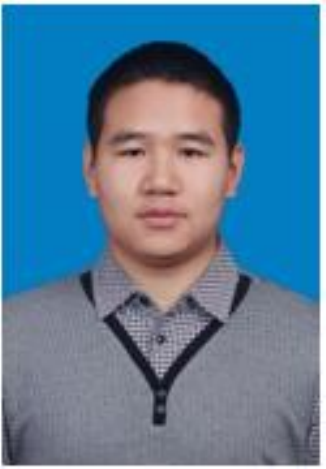}}]{Dayong Shen}
received the bachelor¡¯s degree and master's degree in system engineering from NUDT respectively in 2011 and 2013. He is currently working toward the Ph.D degree in Social Transportation and Social Logistics. His current research interests include intelligent scheduling ,artificial intelligence algorithm and parallel social systems. He has rich experience in designing and implementing parallel logistics system projects.
\end{IEEEbiography}

\begin{IEEEbiography}[{\includegraphics[width=1in,height=1.25in,clip,keepaspectratio]{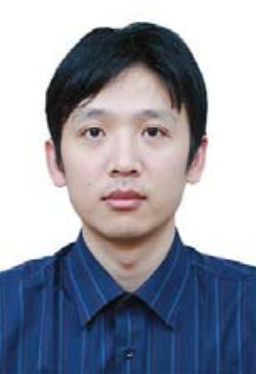}}]{Chen Tian} is an associate professor at State Key Laboratory for Novel Software Technology, Nanjing University, China. He was previously an associate professor at School of Electronics Information and Communications, Huazhong University of Science and Technology, China. Dr. Tian received the BS (2000), MS (2003) and PhD (2008) degrees at Department of Electronics and Information Engineering from Huazhong University of Science and Technology, China. From 2012 to 2013, he was a postdoctoral researcher with the Department of Computer Science, Yale University. His research interests include data center networks, network function virtualization, distributed systems, Internet streaming and urban computing.
\end{IEEEbiography}

\begin{IEEEbiography}[{\includegraphics[width=1in,height=1.25in,clip,keepaspectratio]{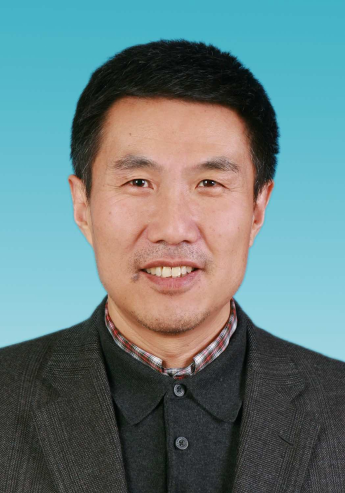}}]{Xiang-Yang Li} is a professor School of Computer Science and Technology, University of Science and Technology of China. He was previously a professor at Computer Science Department at the Illinois Institute of Technology. Dr. Li received MS (2000) and PhD (2001) degree at Department of Computer Science from University of Illinois at Urbana-Champaign, a Bachelor degree at Department of Computer Science and a Bachelor degree at Department of Business Management from Tsinghua University, P.R. China, both in 1995. His research interests include wireless networking, mobile computing, security and privacy, cyber physical systems, and algorithms. He and his students won five best paper awards (IEEE GlobeCom 2015, IEEE HPCCC 2014, ACM MobiCom 2014, COCOON 2001, IEEE HICSS 2001), one best demo award (ACM MobiCom 2012). He published a monograph "Wireless Ad Hoc and Sensor Networks: Theory and Applications". He co-edited several books, including, "Encyclopedia of Algorithms". Dr. Li is an editor of several journals, including IEEE Transaction on Mobile Computing, and IEEE/ACM Transaction on Networking. He has served many international conferences in various capacities, including ACM MobiCom, ACM MobiHoc, IEEE MASS. He is an IEEE Fellow and an ACM Distinguished Scientist.
\end{IEEEbiography}

\begin{IEEEbiography}[{\includegraphics[width=1in,height=1.25in,clip,keepaspectratio]{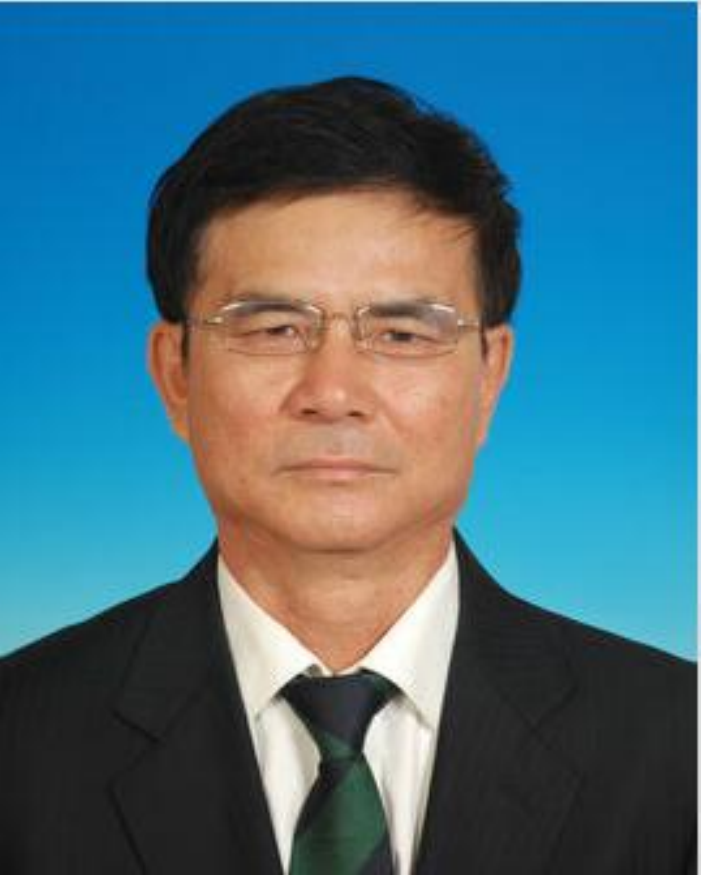}}]{Zhengxi Li}
 is doctoral supervisor, professor and vice president of North China University of Technology. His research interests cover intelligent traffic control and management, control theory and control engineering, electric drive technology.
\end{IEEEbiography}

\end{document}